\documentclass[authoryear]{elsarticle}
\pdfoutput=1

\usepackage{amsmath,amssymb,amsthm}
\usepackage{natbib}
\usepackage{float}
\usepackage{caption}
\usepackage{subcaption}
\usepackage{graphicx}
\usepackage{color}
\usepackage{xspace}
\usepackage[export]{adjustbox}
\usepackage{xcolor}
\usepackage[colorlinks = true,
            linkcolor = black,
            urlcolor  = black,
            citecolor = blue,
            anchorcolor = black]{hyperref}
\usepackage{lineno}
\usepackage{siunitx}
\makeatletter
\def\ps@pprintTitle{%
  \let\@oddhead\@empty
  \let\@evenhead\@empty
  \let\@oddfoot\@empty
  \let\@evenfoot\@oddfoot
}
\makeatother

\DeclareMathOperator*{\argmax}{arg\,max}
\renewcommand{\P}{\mathbf{P}}
\newcommand{\X}{\mathbf{X}}
\newcommand{\F}{\mathbf{F}}
\newcommand{\Z}{\mathbb{Z}}
\newcommand{\E}{\mathbb{E}}
\newcommand{\bnd}{\Gamma}
\renewcommand{\L}{\mathcal{L}}
\newcommand{\NDVI}{\textnormal{NDVI}\xspace}
\newcommand{\BCP}{\textsc{bcp}\xspace}
\newcommand{\BFAST}{\textsc{bfast}\xspace}
\newcommand{\DDM}{\textsc{ddm}\xspace}
\newcommand{\DDMMV}{\textsc{ddm-mv}\xspace}
\newcommand{\DDMBCP}{\DDM-\BCP}
\newcommand{\BCPNDVI}{\textsc{bcp-ndvi}\xspace}
\newcommand{\BCPPC}{\textsc{bcp-pc}\xspace}
\newcommand{\wt}[1]{\widetilde{#1}}
\theoremstyle{remark}

\newcommand*{\rom}[1]{\expandafter\@slowromancap\romannumeral #1@}

\makeatletter
\def\@fnsymbol#1{\ensuremath{\ifcase#1\or *\or \ddagger\or
   \mathsection\or \mathparagraph\or \|\or **\or \dagger\dagger
   \or \ddagger\ddagger \else\@ctrerr\fi}}
\makeatother

\title{Enhancing Environmental Enforcement with Near Real-Time Monitoring: Likelihood-Based Detection of Structural Expansion of Intensive Livestock Farms\tnoteref{t1,t2}}

\author[st]{Ben Chugg\fnref{fn1}}
\ead{benchugg@law.stanford.edu}

\author[st]{Brandon Anderson\fnref{fn1}}
\ead{banderson@law.stanford.edu}

\author[st]{Seiji Eicher}
\ead{seicher@stanford.edu}

\author[st]{Sandy Lee}
\ead{sl269@stanford.edu}

\author[st]{Daniel E. Ho\corref{cor1}}
\ead{dho@law.stanford.edu}

\cortext[cor1]{Corresponding author}
\fntext[fn1]{Joint Contribution.}
\address[st]{Stanford University}
\tnotetext[t1]{This research was supported by Schmidt Futures, Stanford Impact Labs, the Chicago Community Trust, and Stanford's Woods Institute for the Environment. We gratefully acknowledge the Stanford Data Science for Social Good program, in particular contributions from Narasimhan Balasubramanian, Qijia Jiang, and David Kang, and exceptionally helpful comments from Elinor Benami.}
\tnotetext[t2]{Published in the \emph{Journal of Applied Earth Observation and Geoinformation.} \\ \url{https://doi.org/10.1016/j.jag.2021.102463}}

\date{}

\begin{document}

\begin{abstract}
Much environmental enforcement in the United States has historically relied on either self-reported data or physical, resource-intensive, infrequent inspections.  Advances in remote sensing and computer vision, however, have the potential to augment compliance monitoring by detecting early warning signs of noncompliance.  We demonstrate a process for rapid identification of significant structural expansion using Planet's 3m/pixel satellite imagery products and focusing on Concentrated Animal Feeding Operations (CAFOs) in the US as a test case. Unpermitted building expansion has been a particular challenge with CAFOs, which pose significant health and environmental risks. Using new hand-labeled dataset of 145,053 images of 1,513 CAFOs, we combine state-of-the-art building segmentation with a likelihood-based change-point detection model to provide a robust signal of building expansion (AUC = 0.86). A major advantage of this approach is that it can work with higher cadence (daily to weekly), but lower resolution (3m/pixel), satellite imagery than previously used in similar environmental settings.  It is also highly generalizable and thus provides a near real-time monitoring tool to prioritize enforcement resources in other settings where unpermitted construction poses environmental risk, e.g. zoning, habitat modification, or wetland protection.  
\end{abstract}


\maketitle

% \newpage

% \section*{Highlights}
% \begin{itemize}
% \item We develop methods for near real-time monitoring for environmental enforcement. 
% \item Methods are needed particularly for high-cadence, lower-resolution imagery. 
% \item Our method enables changepoint detection in such time series of segmented images. 
% \item We validate the method with intensive livestock farms that pose environmental risk. 
% \item The maximum likelihood approach outperforms state-of-the-art baselines. 
% \end{itemize}

% \newpage    

\section{Introduction}
\label{sec:intro}

The protection of land, air, and water depends critically on the enforcement of environmental laws \citep{gray2011effectiveness}. There is, however, mounting evidence of serious challenges facing environmental regulators~\citep{evansaautomated,us2008concentrated,purdy2010using}. Conventionally, the mainstay of enforcement has consisted of low-frequency visits to permitted facilities (e.g., once every 2-5 years under the U.S. Clean Water Act, the federal law governing water pollution), placing a heavy burden on state and federal agencies in light of growing environmental challenges and budgetary threats. This has lead to increasing interest in leveraging machine learning to augment regulatory capacity~\citep{glicksman2017technological, hino2018machine}.

Academic interest has focused on satellite imagery given rapid progress in the field of computer vision  \citep{gauthier2007integrated,weinstein2018computer} and the dramatic increase in the availability of satellite imagery~\citep{handan2020deep}. In contrast to aerial imagery, which can be hard to obtain and limited to a specific region, optical satellite imagery has become available for many locations at a moderate to low cost. 
The application of deep learning, particularly of convolutional neural networks (CNNs), to such imagery has led to recent breakthroughs in a wide range of regulatory areas including animal tracking  \citep{laradji2020counting, xue2017automatic}, industrial farm detection \citep{handan2020deep, cafo}, monitoring habitat change \citep{evansaautomated}, tracking oil spills \citep{krestenitis2019early,nieto2018two,bianchi2020large}, and deforestation mapping \citep{maretto2020spatio,de2020change}.

Despite these successes and the availability of imagery, a practical barrier in its broader use for regulation stems from the trade-off between temporal and spatial resolution. To date, imagery that is both available to policymakers and at high resolution rarely revisits the same location in the time required for regulatory monitoring. The United States, for instance, has regularly acquired 1m/pixel resolution in the National Agriculture Imagery Program, but this data is only refreshed every few years. Methods that rely on such high spatial resolution imagery are thus ill-suited for augmenting regulatory capacity.  Coarser resolution imagery, however, is increasingly  available at weekly, if not daily, revisit times (see, e.g., Sentinel-2 Imagery).  Such data could aid, for instance, in detecting zoning violations, building construction, and wetlands destruction, but requires a method for detecting changes in time series that align with the spatial and temporal resolution of what is more broadly available today.

In this work, we present a technique intended to address this important gap by statistically leveraging repeated observations of medium-resolution (3m/pixel), but high-cadence (weekly), imagery to augment environmental enforcement.\footnote{We acknowledge that describing 3m/pixel data as ``medium'' resolution is relative. We describe it as such because existing work in similar environmental settings (e.g., detecting CAFOs) has relied on imagery with at least 1m/pixel resolution.} The method builds on any given segmentation approach and requires only simple (and generalizable) assumptions about object permanency.

To demonstrate the technique, we focus on the task of monitoring the capacity of Concentrated Animal Feeding Operations (CAFOs), which can generate serious environmental and health consequences (see Section~\ref{sec:cafos}).  We first train an image segmentation model to differentiate facilities from background, and then test for change points within a time series of satellite images.

Our contributions are: 
\begin{enumerate}
    \item A new time series dataset of 145,053 hand-validated segmented satellite images (GeoTiffs) for 1,513 CAFO facilities from 
    January 1 to December 31, 2019;\footnote{Our underlying imagery comes from Planet Labs's PlanetScope Daily Imagery.}
    \item A maximum likelihood model that takes as inputs a time series of segmented images and provides a test statistic that captures the likelihood that a facility expanded; 
    \item A demonstration that this model can detect the expansion of CAFO facilities in Indiana with high-cadence satellite imagery;
    \item A comparison to show that this model provides substantial improvement relative to changepoint detection baselines.
\end{enumerate}

Replication code and data are available at:  \url{https://github.com/reglab/building_expansion}. 

The remainder of the paper is structured as follows. Section~\ref{sec:cafos} provides  background on CAFOs, the substantial environmental impact of intensive livestock farming, and the significant gap in environmental monitoring. Section~\ref{sec:related_work} discusses related work in detecting structural change in satellite imagery. Sections~\ref{sec:data} and~\ref{sec:methods} provide background on the data and the details of our proposed algorithm. Section~\ref{sec:performance_metrics} presents the performance we use to analyze the results, and Section~\ref{sec:results} provides results. Section~\ref{sec:discussion} discusses implications, interprets the differences in results, and offers future directions for research.

\section{Background on CAFOs}
\label{sec:cafos}

\begin{figure}[t]
    \centering
    \includegraphics[scale=0.2]{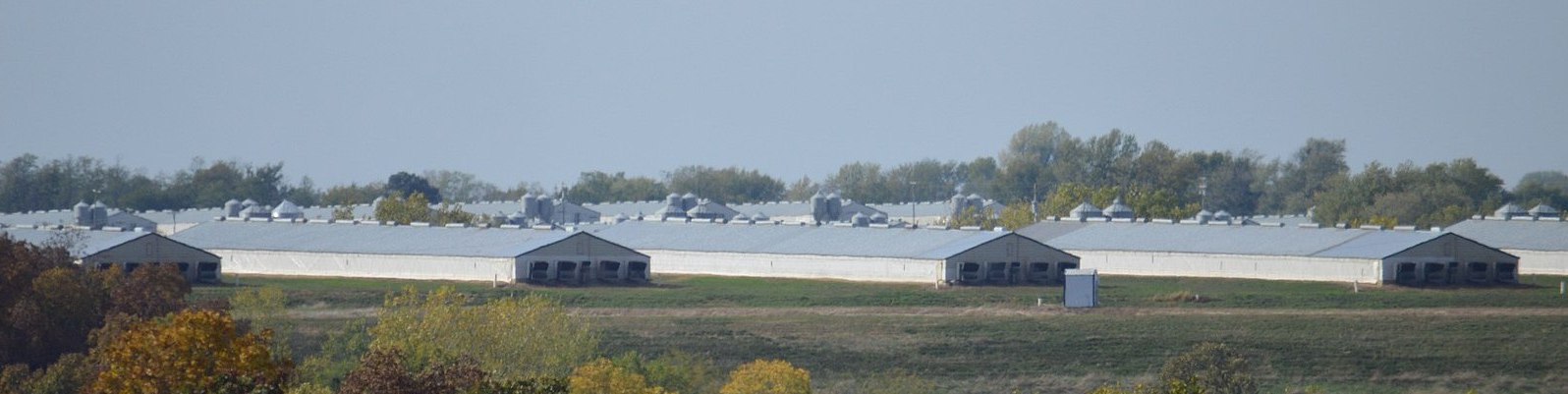}
    \caption{CAFO in Unionville, Missouri housing large volumes of livestock. Each of the four buildings in the foreground are CAFO sheds. Source: Socially Responsible Agriculture Project.  } 
    \label{fig:cafo}
\end{figure}

Large-scale industrial farming is increasingly responsible for livestock production in the United States~\citep{macdonald2009transformation, macdonald2018threedecadesconsolidation,hribar2010understanding}. CAFOs, responsible for raising large numbers of animals at high densities, represent a  prominent feature of this transformation. They have been estimated to produce anywhere from 40\% to 99\% of all livestock in the United States \citep{copeland2010animal,sherman2008,anthis_us_2019}, with uncertainty stemming from lack of regulatory monitoring. Figure~\ref{fig:cafo} provides an example of a large CAFO, with distinctive large sheds that confine animals.  Under federal law, a large CAFO, for instance, contains 125,000 or more heads of poultry or 2,500 or more heads of hog or 1000 beef cows.\footnote{See EPA, Regulatory Definitions of Large CAFOs, Medium CAFO, and Small CAFOs, \url{https://www.epa.gov/sites/production/files/2015-08/documents/sector_table.pdf}.} 

CAFOs pose a variety of social and environmental risks with limited and infrequent oversight. For example, a large CAFO can produce 
some 1.6M tons of waste a year -- which is equivalent to a large (1M resident) city -- but is not subject to the same regulation as human wastewater~\citep{us2008concentrated}. Such large levels of animal waste are thought to be serious contributors to water contamination, due to potentially poorly-constructed manure storage pits or run-off from the application of waste to fields~\citep{burkholder2007impacts,sherman2008}, as well as air pollution \citep{cafo_impact, hribar2010understanding}. Consequences of this contamination include the degradation of aquatic systems, harmful algal blooms, and nitrate contamination of drinking water~\citep{copeland2010animal,hribar2010understanding}. These health and environmental risks can are frequently borne by poorer and minority communities, where such facilities are disproportionately sited~\citep{nicole2013cafos, son2021distribution}.

The regulation of CAFOs has proven challenging. As noted by the Government Accountability Office (GAO), the ``EPA does not have comprehensive, accurate information on the number of permitted CAFOs nationwide. As a result, EPA does not have the information it needs to effectively regulate these CAFOs''~\citep{us2008concentrated}. There is substantial evidence that many CAFOs often undergo unpermitted construction and expansion, in the process evading environmental review~\citep{Howard2019Illinois, marshall, guay, merced}. While we draw upon building expansion in CAFOs as a focused case where evasion is acute, this type of regulatory avoidance also exists in other areas of environmental law, where limited monitoring systems and the reliance on self-reporting can impede regulatory efforts~\citep{OIG, purdy2010using}.
Given the potential social and environmental impact, it is critical to develop methods for closer to real-time monitoring.

\section{Related Work}
\label{sec:related_work}

% Prior work that uses machine learning techniques on satellite data to detect changes in building extent often focus on urban areas. For example, using the ZY3 constellation of satellites that simultaneously image an area from multiple viewing angles, \citet{huang2020automatic} extract information about building height to detect newly constructed buildings in Beijing and Shanghai. \citet{malpica2013change} use support vector machines on a combination of satellite imagery and LiDAR to detect changes. \citet{chen2019changenet} use generative adversarial networks to identify pixel-wise differences in pairs of images. \citet{leichtle2017unsupervised} apply PCA and $k$-means clustering to very fine resolution imagery (50cm/pixel). Synthetic aperture radar (SAR) data have also frequently been used to detect urban expansion (e.g.,~\citep{ban2017eo4urban, del2008monitoring, gamba2006change, ban2012multitemporal, marin2014building}). Recent work has also examined the use of change detection techniques that exploit data from optical and RADAR imagery, finding higher accuracy/precision with flood events relying on both sets of data  \citep{liu2017change, luppino2017clustering}. 

For general change detection techniques on lower-cadence imagery, we refer the reader to the surveys of \citet{zhu2017change} and \citet{namoano2019online}.  Previous work focusing on longer time series is typically concerned with larger scale changes than the addition of buildings, e.g., urban expansion~\citep{wan2019mapping}, changing land cover~\citep{zhu2014continuous}, forest disturbance~\citep{kennedy2010detecting, huang2010automated}, or vegetation~\citep{verbesselt2010phenological, browning2017breaks}. A common approach for longer sequences of imagery is to reduce each image to a metric --- a vegetation or drought index for instance --- and apply more traditional changepoint detection techniques to the resulting time series. The Normalized Difference Vegetation index (\NDVI) is one popular metric, used for example by both \citet{wan2019mapping} to detect urban change in Landsat imagery and by the Breaks for Additive Season Trend (\BFAST) algorithm~\citep{verbesselt2010detecting}.\footnote{\NDVI gives a pixel-wise value (between -1 and 1) corresponding to the amount of greenery in the pixel, and can be reduced to a single value for the image by averaging.} 
\cite{setiawan2016simple} proposes a median moving window and linear interpolation to reduce noise in the \NDVI time series. \cite{ye2021near} propose a state space model with a Kalman filter and take a time series approach to detect forest disturbance. 

Another strand of research uses regression methods for anomaly detection. The influential approach by \cite{zhu2014continuous} develops a Continuous Change Detection and Classification algorithm, using ordinary least squares time series to classify  changes at the pixel level. \citet{koltunov2009image} describes the Dynamic Detection Model (\DDM), which combines a series of basis images into a prediction image. This prediction image is then compared with the true test image for anomaly detection.  The \DDM and related anomaly detection approaches have inspired a wide range of applications  \citep[e.g.,][]{tang2020can, koltunov2020edart}. Another approach is by \citet{fytsilis2016methodology}, which expands anomalous pixels into homogeneous regions and calculates a difference metric between two images based on the maximum spatial correlation of the shape with each image as a whole. 

% Add Para deep learning techniques --> vs. lightweight approach when ground truth about is not available. 
When sufficient data are available, deep learning based detection methods have also proven useful. \citet{varghese2018changenet} develop ChangeNet, a CNN to detect changes between pairs of images.  
\citet{peng2019end} develop an end-to-end (as opposed to image-by-image or pixel-by-pixel) change detection method using UNet++, and \citet{sefrin2021deep} conjoin two deep learning architectures to detect land cover change. 
While promising, such deep learning approaches cannot easily be applied in our setting due to the large number of free parameters and scarce positive examples. 

The most closely related work is that of \citet{robinson2021temporal}, who proposed a semi-supervised algorithm for retroactively determining construction dates using high-resolution satellite imagery.  While the approach is complementary, it presupposes a set of final structure footprints, within which it looks backwards in time for localized spectral shifts that represent construction events.  Having these footprints, however, would imply that the expanded or new structures were already detected at a reasonable significance. Instead, the  premise of our work here is that this is a challenging task at lower resolutions.  Instead of focusing on the detection of individual buildings, our method tests the broader hypothesis that the distribution of pixels classified as a structure undergoes a discrete change over time.  This allows us to draw more reliable inferences using a statistical framework for building expansion in  high-cadence, lower-resolution imagery.

\begin{figure}[t]
    \centering
    \includegraphics[scale=0.44,center]{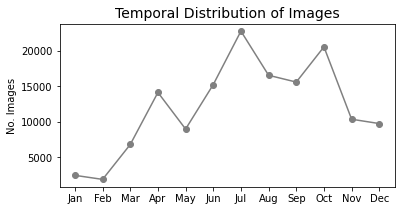}
    \caption{Total number of images in each month across all locations.} 
    \label{fig:image_counts_and_ndvi}
\end{figure}

In sum, while much of the above work has demonstrated that detecting changes in local building footprints is possible, we are unaware of work that has attempted to provide near real-time (weekly), actionable insights for regulatory enforcement. Below, we compare existing approaches, showing substantial gains to our likelihood approach tuned to this setting. 

\section{Data}
\label{sec:data}

\subsection{Imagery}

Our dataset consists of 1,516 ground truth Indiana CAFO coordinates validated by the Environmental Working Group (EWG). At each location, we queried the Planet Labs API for 4-band (red, green, blue, and near-infrared) imagery between January 1, 2019 and December 31, 2019, requiring that each be at least 95\% clear (e.g., of cloud cover and cloud shadows). This yielded roughly 80 to 130 images per location, with 175,736 images total. Due to the cloud cover filter, images were less frequent in January, February, and December, and most frequent in the summer (see Figure~\ref{fig:image_counts_and_ndvi}). 
% The API call clips the 24 km by 7 km scenes to a 600m by 600m area of interest (AOI) specified by a GeoJSON polygon.
We clipped each resulting scene to a centered 600m$\times$600m (200$\times$200 pixel) area, and discarded those with over 15\% missing pixel values (missing pixels occur if the image was on the edge of the area being considered). 
% \footnote{Such missingness occurs if the area of interest falls on the edge or corner of a scene band.}
Overall, post-processing removed just over 30,000 images, leaving 145,053 corresponding to 1,513 final locations.

\begin{figure}
    \begin{minipage}{0.2\textwidth}
    \textbf{Facility A}\\
    \includegraphics[scale=0.24,right]{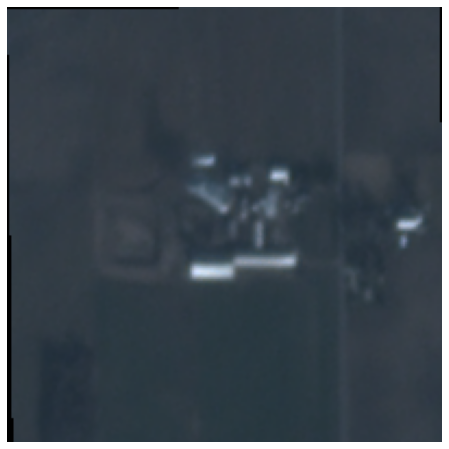}
    \subcaption*{(1) Before}
    \end{minipage}
    \hspace{-0.3cm}
    \begin{minipage}{0.2\textwidth}
    \textcolor{white}{\textbf{Facility A}}\\
\includegraphics[scale=0.24,left]{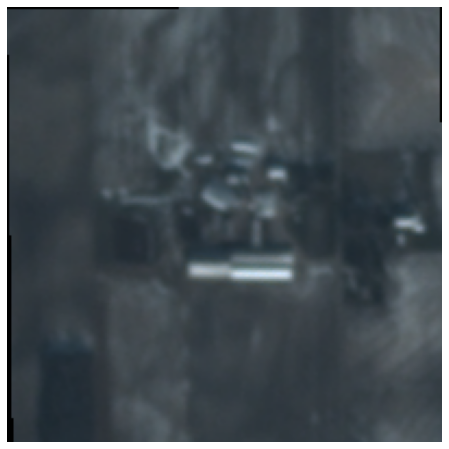}
    \subcaption*{(2) After}
    \end{minipage}
    \hspace{2.2cm}
    \hspace*{0.2cm}
    \begin{minipage}{0.2\textwidth}
    \textbf{Facility B}\\
    \includegraphics[scale=0.24,right]{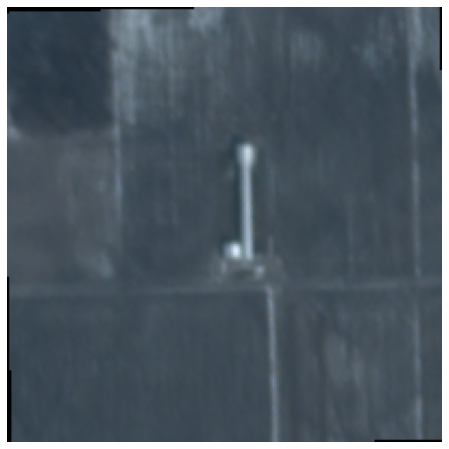}
    \subcaption*{(3) Before}
    \end{minipage}
    \hspace{-0.3cm}
    \begin{minipage}{0.2\textwidth}
    \textcolor{white}{\textbf{Facility B}}\\
    \includegraphics[scale=0.24,left]{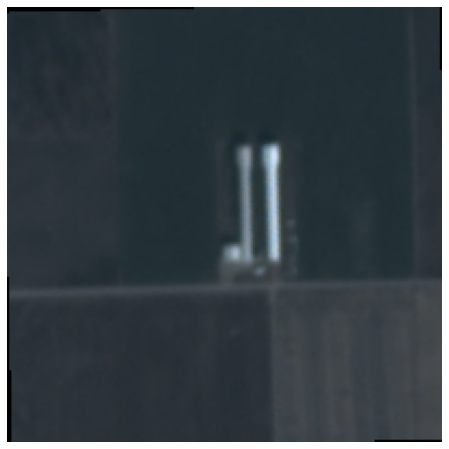}
    \subcaption*{(4) After}
    \end{minipage}
    \caption{Two examples of expanded CAFO facilities. Both examples exhibit environmental factors which can complicate the segmentation---cloud cover in panel (2) and snow in the panel (3). } 
    \label{fig:expansion}
\end{figure}

Figure~\ref{fig:expansion} illustrates both the resolution of the imagery and gives two examples of CAFO expansion. We note that these images are of lower spatial resolution (3m/pixel) than previously used for CAFOs and other similar settings, which poses the main inferential challenge for real-time monitoring.  To the human eye, the construction of barns is visible (i.e., a third rectangular roof to the south of facilities in the left panel and a second rectangular roof to the east of the single barn in the right panel). The images, however, also illustrate some of the complications for computer-based object segmentation, such as the cloud cover in panel (2) and the snow in panel (3). 

\subsection{Building Footprints}

Many applications may be able to take advantage of pre-existing segmentation models. Conventional building segmentation models, however, have focused in large part on urban areas, which may result in performance degradation in rural areas. To provide building footprints that are tailored to CAFOs --- and not other types of structures --- we train our own segmentation model (see Section~\ref{subsec:segment}).  This approach required ground truth building footprints corresponding to known CAFO locations.
% Next, we generate ground truth labels for the images so that they can be used as training data for the image segmentation model. 
We procured such information from (1) the Microsoft Building Footprint dataset~\cite{ms}, and (2) a United States Geological Survey of poultry barns in the DelMarVa (Delaware - Maryland - Virginia) peninsula~\citep{delmarva}. %
%
%
% Beginning with the Microsoft Building Footprint dataset, the results of Microsoft's U.S.-wide building segmentation project in roughly 2017/2018 as ground truth, we rasterized the Indiana shapefile against each downloaded image, resulting in a $200\times 200$ binary mask, with 0 indicating a background pixel and 1 indicating a CAFO pixel.%
%
Because the Microsoft dataset includes all building types, we removed those footprints with area less than 450 $\text{m}^2$ that are below the threshold of a reasonable size for large CAFO operations. When combined with the proximity to a known CAFO location, this filter left us with a fairly pure training set. The most significant remaining source of noise stemmed from date mismatch between the Microsoft and EWG labels.  For instance, a CAFO may have been constructed and tagged \emph{after} the building footprints were generated.  Similarly, the DelMarVa data was collected in 2016-2017 -- these barns may have been demolished or moved between the time of data collection and the acquisition dates of our Planet imagery datasets (2019). As far as we are aware, such changes are rare, but they would reduce the accuracy of our image segmentation and hence attenuate results of subsequent changepoint detection. 
%\ba{should include here the number of locations from each source} Working on it. 

\subsection{Expansion Labels}
\label{sec:expansion_labels}

Each image time series corresponding to our CAFO locations was hand-labeled by our research team to indicate whether a facility expanded during the observation window (1/2019--12/2019). This was done by comparing the first and last available images in the time series, with the review of intermediate images if any structural changes were observed. In total, we found 22 sites that had constructed at least one new CAFO building, 1,414 with no evidence of new construction, and 77 with either a no-longer-existent CAFO or indeterminate expansion.  
It is worth noting that despite the strong class imbalance, the sheer number of farm operations in the US --- just shy of one million according to the National Agricultural Census~\citep{agcensus}, excluding aquaculture --- implies that there are potentially on the order of 10,000 such expansion events every year.

\section{Methods}
\label{sec:methods}

Our procedure for detecting structural expansion can be broken down into three  steps. In the first, we apply a segmentation model~\citep{ronneberger_u-net_2015} to each image in order to obtain pixel-level probabilities of class membership, with higher values indicating that the pixel is more likely to belong to a CAFO shed. 
We emphasize that our approach is independent from the choice of segmentation model as long as it yields such probabilities.\footnote{A notable alternative for an image segmentation approach would be DeepLabv3+~\citep{chen2018encoder}, which has shown small performance gains of roughly 1\% in accuracy relative to U-Net ~\citep{ahmed2020comparison, cai2021comparative, jiwani2021semantic}. In~\ref{app:segment}, we conduct an ablation study showing that performance of MLE is comparable across reasonable ranges of segmentation accuracy.}
In the second step, we fit a time-dependent building footprint model using maximum likelihood estimation (MLE) to best match the class probabilities. 
Finally, we compare this time-dependent model to a restricted model fit under the hypothesis of no building expansion.  
A test statistic is then computed comparing the two fits.  
Intuitively, the more dissimilar the two fits, the more likely there was expansion. 

\subsection{Segmentation}
\label{subsec:segment}

In order to take advantage of both the spectral and spatial signatures of a CAFO structure, we employ a U-Net architecture \citep{unet}. We trained the model on about 2,611 labeled images, composed of 1,176 EWG location images with post-processed Microsoft building footprint labels and 1,435 images of DelMarVa peninsula poultry houses with labels from their shapefile. DelMarVa data, in particular, appeared to boost the performance of the segmentation model. We trained on one image per location, with an even split throughout each month of the year to regularize the model against seasonality. 

The model was trained with a batch size of 8, using the Adam optimizer with  learning rate 5e-4, and weight decay 1e-7. The training data was split into 70\% train, 15\% validation, and 15\% test sets. We performed random flips and rotations on the training set for data augmentation. The loss function converged after 20 epochs. Our loss function was binary cross-entropy with logits, and we weighted CAFO to background loss 30:1 to compensate for the class imbalance. 
%Figure \ref{fig:seg_stats} shows the final results of the model on the validation set. With a .864 recall and .4818 precision, it was prone to overpredicting the positive class. 
The final results on the validation set had a recall of 0.864, a precision of 0.482, and an Intersection over Union (IoU) of 0.448. 

\begin{figure}
    \centering
    \begin{minipage}{0.49\textwidth}
    \includegraphics[scale=0.3,center]{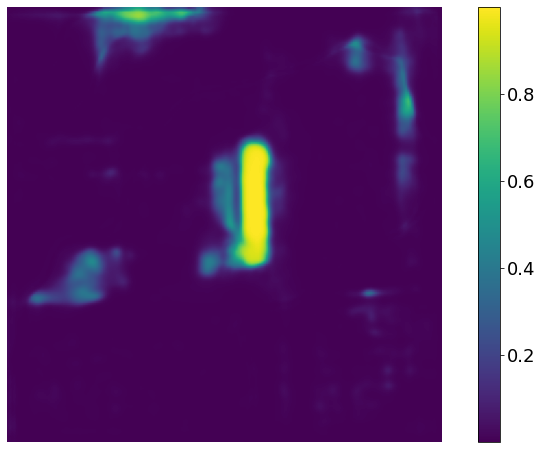}
    \subcaption*{Before}
    \end{minipage}
    \begin{minipage}{0.49\textwidth}
    \includegraphics[scale=0.3,center]{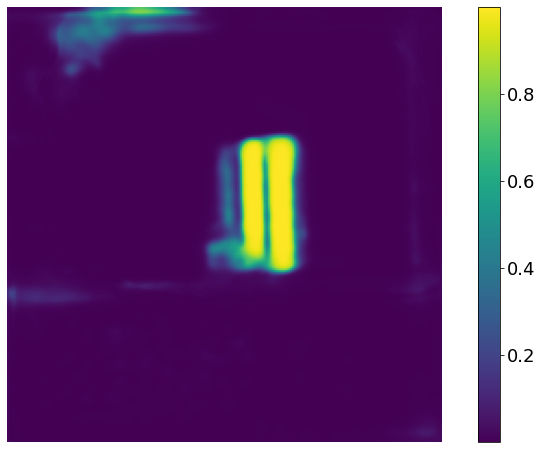}
    \subcaption*{After}
    \end{minipage}
    \caption{Pixel-level class probabilities determined by the segmentation model of the rightmost example of Figure~\ref{fig:expansion}. Higher values indicate a higher likelihood of being part of a CAFO.   }
    \label{fig:seg_probs}
\end{figure}

Our change detection algorithm (explained below) does not take as input the final segmentation, but rather pixel-level class probabilities for each image (CAFO vs. no-CAFO). Instead of passing along the probabilities directly from the U-Net, however,\footnote{Our U-Net implementation did not provide class probabilities directly, but rather class \emph{confidences} in the range [-7,7]. We applied the sigmoid function to these values in order to obtain class probabilities.} a final post-processing step was applied: A $k\times k$ smoothing kernel ($k=3$) was applied to each image, whereby the probability of each pixel was average over the $k^2$ nearest pixels. Smoothing was applied in order to remove small artifacts generated by the segmentation model. For each location, the corresponding series of probability maps was then converted into a tensor in order to apply MLE as described next. Figure~\ref{fig:seg_probs} illustrates the probabilities generated by the segmentation model. 

% Finally, we applied post-processing on the predicted masks using the openCV package. We culled any contours on the masks that were of pixel area less than 50, and fit rotated bounding rectangles to each contour that remained in order to smooth out noisy predictions. We took the final result as the image's building segmentation mask.

\subsection{Footprint Modeling and Change Detection}
\label{sec:change_detection}

Once the imagery has been segmented, we can test each CAFO location for significant changes in building footprint over a series of time-sequenced snapshots.  For clarity of demonstration in our test case we make the following simplifying assumptions:

\begin{enumerate}
    \item CAFO sheds are constructed and not removed.
    \item The time required to construct a CAFO shed is short compared to the length of the period considered.
    \item Only one expansion event can take place during the period considered.
\end{enumerate}

These assumptions may be relaxed, affecting the parameterization of the likelihood approach. Formally, let $\{\X^1,\dots,\X^T\}$ with $\X^t\in \Z_{\geq 0}^{w\times h\times 3}$ be the series of (3-band) images for some fixed location at each timestep $t\in\{1,\dots,T\}$. Here, $w=200$ is the width of the image and $h=200$ its height in pixels. For each image $\X^t$ the U-Net produces a matrix of class-membership probabilities $\P^t\in[0,1]^{w\times h}$, where $p^t_{ij}=\P^t_{ij}$ is the probability that pixel $(i,j)$ belongs to a CAFO shed.

\begin{figure}
\centering
    \begin{minipage}{0.46\textwidth}
    \includegraphics[scale=0.24,center]{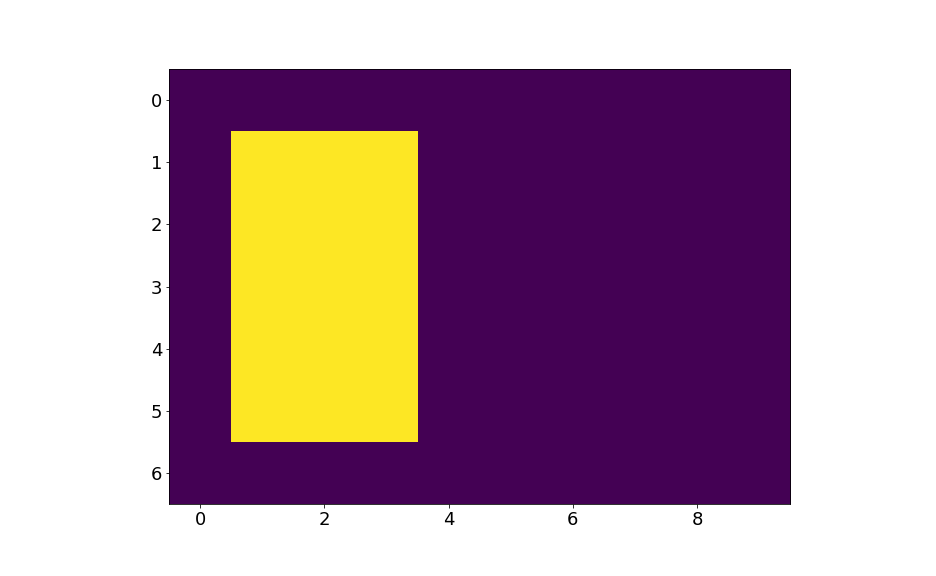}
    \end{minipage}
    \begin{minipage}{0.46\textwidth}
    \includegraphics[scale=0.24,center]{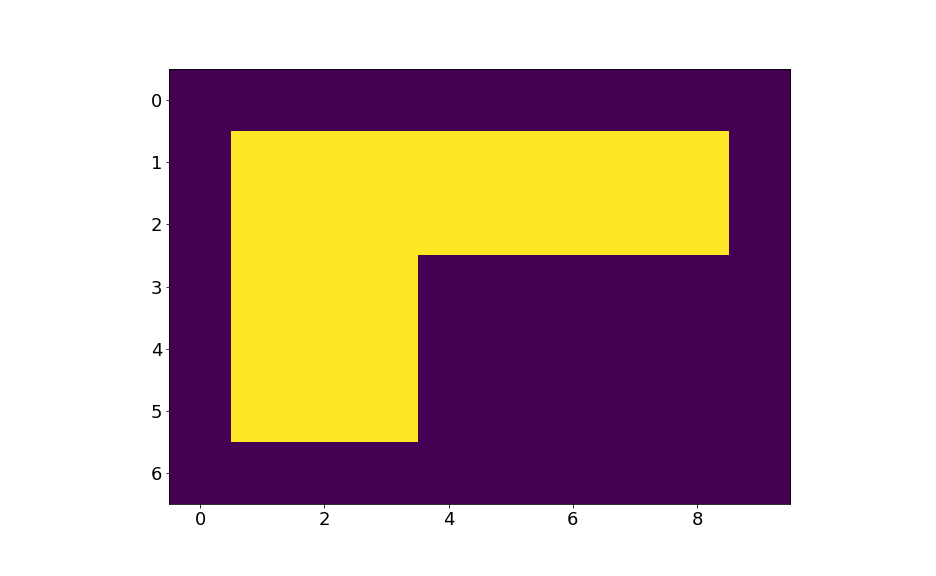}
    \end{minipage} \\
    \begin{minipage}{0.49\textwidth}
    \centering
    $
    \F^0=
    \tiny
    \begin{bmatrix}
    0 & 0 & 0 & 0 & 0 & 0 & 0  & 0 & 0 & 0   \\
    0 & 1 & 1 & 1 & 0 & 0 & 0  & 0 & 0 & 0   \\
    0 & 1 & 1 & 1 & 0 & 0 & 0  & 0 & 0 & 0   \\
    0 & 1 & 1 & 1 & 0 & 0 & 0  & 0 & 0 & 0   \\
    0 & 1 & 1 & 1 & 0 & 0 & 0  & 0 & 0 & 0   \\
    0 & 1 & 1 & 1 & 0 & 0 & 0  & 0 & 0 & 0   \\
    0 & 0 & 0 & 0 & 0 & 0 & 0  & 0 & 0 & 0   
    \end{bmatrix}
    $
    \end{minipage}
    \begin{minipage}{0.49\textwidth}
    % \raggedright
    $
    \F^+=
    \tiny 
    \begin{bmatrix}
    0 & 0 & 0 & 0 & 0 & 0 & 0  & 0 & 0 & 0   \\
    0 & 0 & 0 & 0 & 1 & 1 & 1  & 1 & 1 & 0   \\
    0 & 0 & 0 & 0 & 1 & 1 & 1  & 1 & 1 & 0   \\
    0 & 0 & 0 & 0 & 0 & 0 & 0  & 0 & 0 & 0   \\
    0 & 0 & 0 & 0 & 0 & 0 & 0  & 0 & 0 & 0   \\
    0 & 0 & 0 & 0 & 0 & 0 & 0  & 0 & 0 & 0   \\
    0 & 0 & 0 & 0 & 0 & 0 & 0  & 0 & 0 & 0   
    \end{bmatrix}
    $
    \end{minipage}
    \caption{Idealized CAFO footprints before and after expansion (top left and right, respectively). Yellow (light) squares represent the CAFO shed. The entries with a 1 in $\F^0$ (rows 1--5, columns 1--3 assuming a zero-index) represent yellow pixels in the footprint prior to expansion. The entries with 1 in $\F^+$ represent the pixels of the added building (rows 1--2, columns 4--8).   }
    \label{fig:footprint_matrix}
\end{figure}

Let $\F^0\in\{0,1\}^{w\times h}$ be the \emph{footprint} of the CAFO shed prior to any expansion. That is, $\F^0_{i,j}=1$ if pixel $(i,j)$ belongs to a CAFO shed, and 0 otherwise. Let $\F^+\in\{0,1\}^{w\times h}$ define the pixels which belong to a CAFO shed \emph{after} expansion, but did not before. Thus, $\F^0+\F^+$ defines which pixels belong to the expanded shed. If there was no expansion then $\F^+$ is the all-zeroes matrix. Figure~\ref{fig:footprint_matrix} illustrates how $\F^0$ and $\F^+$ correspond to an expanded shed.  

Using $\{\P^1,\dots,\P^T\}$, the goal of the expansion model is to determine $\F^0$, $\F^+$ and at what timestep (if any) the transition occurs. To this end, we define a function which, given the building footprints and transition time $t^*$, captures the transition:
\begin{equation}
\label{eq:transition}
    Z_{\F^0,\F^+,t^*,\alpha}(t) = \F^0(1 - S_\alpha(t-t^*)) + \F^+ S_\alpha(t-t^*),
\end{equation}
where $S_\alpha(x) = \alpha/ (1 + e^{-x})$ is the sigmoid function and $t^*$ is the timestep at which the CAFO shed expands. Here, $\alpha$ controls the transition speed: bigger values of $\alpha$ imply a longer building period. 
For notational convenience, $Z_{\F^0,\F^+,t^*,\alpha}(t)$ will be written as $Z(t)$ or simply $Z^t$. Note that for $t<t^*$, $Z(t)\approx \F^0$ (the original CAFO shed), and for $t>t^*$, $Z(t)\approx\F^0 + \F^+$ (the expanded shed). 

In what follows, we use maximum likelihood estimation (MLE) to estimate the parameters of $Z$ which are best described by the class probabilities $\P^1,\dots,\P^T$. This is performed twice: once with no restrictions on the parameters, and once with the restriction $\F^0=\mathbf{0}$ (enforcing no expansion).

% With these restrictions, it is possible to express an expanding CAFO footprint model as a tensor $F$ that smoothly transitions between $C^0_{ijk}$, with spatial dimensions $i,j$ and binary class dimension $k$, into an augmented footprint $C^1_{ijk}$ around the $z_0$th frame of the sequence:

% \begin{equation}
%     F_{ijkz}(z_0, C^0_{ijk}, C^1_{ijk}, \alpha) = C^0_{ijk}[1-S(\alpha(z-z_0))] + C^1_{ijk} S(\alpha(z-z_0))
% \end{equation}

% where $S$ is the sigmoid function, and $\alpha$ controls the speed of transition between the two building layouts.  The function forms a null hypothesis (no expansion) by setting either $z_0\le0$ or $\mathbf{C}^1=\mathbf{C}^0$.

We judge the fitness of a given set of parameters by forming a likelihood comparing $Z$ to the pixel-level class probabilities yielded by the segmentation model.  In other words, for each image frame the model will predict a class membership vector for each pixel.  We take the corresponding class probability vector from the segmentation model $p_{ij}^t$ and form a likelihood with their product.  Functionally, this takes the form:

\begin{equation}
    \mathcal{L}_t = \prod_{i,j} p_{ij}^t\cdot Z^t_{ij}.
\end{equation}

Taking the product over all frames in the time sequence to get the likelihood $\L=\prod_t\L_t$, the log-likelihood becomes: 

\begin{equation}
\label{eq:log_likelihood}
    \log\mathcal{L} = \sum_{i,j,t} \text{log}(p_{ij}^t\cdot Z^t_{ij}).
\end{equation}

Observing that $Z^t$ is differentiable due to the sigmoid function, we adjust its parameters to maximize Equation~\eqref{eq:log_likelihood}. For the the unrestricted model, we solve: 
\begin{equation*}
    \wt{\F^0},\wt{\F^+},\wt{t}^*=\argmax_{\F^0,\F^+,t^*} \log \mathcal{L}(\F^0,\F^+,t^*|\P^1,\dots,\P^T).
\end{equation*}
For the static model, we enforce no expansion and solve: 
\begin{equation*}
    \overline{\F^0},\overline{t}^*=\argmax_{\F^0,t^*} \log \mathcal{L}(\F^0,\mathbf{0},t^*|\P^1,\dots,\P^T).
\end{equation*}

From here, we form a test statistic---the delta log-likelihood---comparing the two fits: 

\begin{equation}
\label{eq:ts}
    TS = \log \mathcal{L}(\wt{\F^0},\wt{\F^+},\wt{t}^*|\P^1,\dots,\P^T) - \log \mathcal{L}_0(\overline{\F^0},\F^+=\mathbf{0},\overline{t}^*|\P^1,\dots,\P^T).
\end{equation}

This statistic enables us to assess whether there is a difference in distributions between the images with and without expansion. Intuitively, we expect that the locations exhibiting expansion will larger test statistic (in absolute value) than those that do not, because those without expansion should be well modeled with $\F^+=\bf{0}$. 
In practice, plotting the values of the test statistic for facilities known not to have expanded gives us a null distribution against we can compare the values of other facilities. The further a facility deviates from this null distribution, the more likely it is (according to the model) to have expanded.  

\subsection{Baselines}
\label{sec:baselines}

We compare our method against several baseline methods. Two use Bayesian changepoint detection (\BCP), a recommended method when analyzing univariate timeseries \citep{van2020evaluation}. We apply \BCP to both the sequence of \NDVI values obtained from the images over time (see Section~\ref{sec:related_work}), and to the number of pixels identified as belonging to CAFOs by the segmentation model in each image. We refer to these as \BCPNDVI and \BCPPC, respectively.
Another baseline is the Breaks For Additive Seasonal and Trend (\BFAST) algorithm by \citet{verbesselt2010detecting}. As \BFAST was originally designed to examine vegetation response, we apply it to the timeseries of \NDVI values obtained from the images. 

Our final baseline is based on the Dynamic Detection Model (\DDM). \DDM is a localized anomaly detection method, contrasting each pixel/band value at time $t$ with a predicted value generated by an optimized combination of $m$ prior observations. Specifically, let $\bnd^t_{ijk}$ be a pixel value within a spectral band $k$ of image $\X^t$. A linear \DDM fits coefficients $\gamma(t)\equiv\{\gamma_0(t),\dots,\gamma_m(t)\}$ to form an estimator
\begin{equation}
    \widehat{\bnd}^t_{ijk} = \gamma_0(t) + \sum_{\tau=1}^m \gamma_\tau(t) \bnd^{t-\tau}_{ijk}
\end{equation}

that minimizes the RMSE $\sigma_k(t)$ relative to the true value $\bnd^t_{ijk}$. Note the explicit dependence on $t$, which indicates that each set of $\gamma$ is tied to an absolute time and can therefore be used to describe dynamics such as seasonality.\footnote{To account for finite cadence and gaps in observation, we fit $\gamma$ on aggregated imagery at a 1-week resolution.} The localized anomaly score $Z^t_{ijk}$ is then generated by

\begin{equation}
    Z^t_{ijk} = \frac{\bnd^t_{ijk} - \widehat{\bnd}^t_{ijk}}{\sigma_k(t)}.
\end{equation}

To provide an image-level anomaly score $Z^t$ comparable to our other methods, we simply sum $Z^t_{ijk}$ over pixels and the three visible bands $k\in\{R,G,B\}$, leveraging the fact that true expansion events should involve multiple simultaneous pixel changes. From here, we employ two techniques to determine the probability of an expansion event taking place during a given period at this location. The first is to check if any score $Z^t$ is above a predefined threshold. We call this \DDM-\textsc{Max Value}, or \DDMMV. The second is to apply \BCP to the time-series $\{Z^t\}$, and similarly classify the series as exhibiting an expansion event whenever \BCP returns a probability over a given threshold.  We call this \DDMBCP. When reporting the accuracies for both methods, we scan over the set of all thresholds and report the maximum value. 

It is worth noting that none of the baseline methods were tailored specifically for the purpose of detecting discrete image-level changes in an ongoing series.
Limitations of these approaches reflect the lack of research in this setting, not a general  drawback to the methods. \BFAST, for instance, was designed for timeseries over recurring seasons and large scale changes. And while sensitive to the initial appearance, \DDM does not inherently make use of subsequent observations that reinforce the hypothesis of a new structure.

\section{Performance Metrics}
\label{sec:performance_metrics}
We evaluate our proposed method in addition to three baseline methods. The evaluation uses a combination of metrics, several traditional in machine learning, and several designed to capture resource efficiency.
All four methods assign a ``confidence'' of expansion to each location: a posterior probability for \BCP, a magnitude for \BFAST, and the test statistic (Equation~\ref{eq:ts}) for MLE. 
This is important from a resource allocation perspective, as we would like to know not only which facilities the model think expanded, but which are most likely to have done so. It allows enables us to examine the performance of the various methods as a function of their confidence. 

\paragraph{Balanced Accuracy \& F1-score} Accuracy and F1-score are traditional performance metrics in machine learning. Due to the severe class imbalance, we report balanced accuracy as opposed to overall accuracy. Balanced accuracy is the mean of sensitivity and specificity, and the F1-score is the harmonic mean of precision and sensitivity. Accuracy and F1-score can be calculated using any given confidence as a threshold: Locations with an assigned confidence above the threshold are predicted as expansions. For each algorithm, we report the report the maximum balanced accuracy and F1-score achieved over all confidence values.  

\paragraph{ROC \& AUC} Receiver Operating Characteristic (ROC) Curves and the corresponding area underneath the curve (AUC) are another common metric in machine learning. The ROC Curve plots the false positive rate against the true positive rate as the threshold (in this case, the confidence) of the classifier is varied. The larger the area under the curve the better the overall performance, with an area of one being optimal. In addition to the area underneath the ROC curve, we also report the area under the precision-recall curve (PR-AUC).

\paragraph{Confidence-Size Correlation} Ideally, higher model confidence in an expansion is correlated with a larger expansion. If so, sorting by  confidence would allow enforcement agencies to focus on the largest expansions first. Moreover, a high correlation suggests that the model is picking up on the features indicative of an expansion. We thus examine the Pearson correlation between the size of the expansion and each model's confidence.

\paragraph{Resource Expenditure} Finally, we examine to what extent the four methods save costs compared to a random search --- a reasonable baseline given extensive reliance on random field visits --- for building expansions. Suppose that examining a facility which did not expand has a cost of 1, while correctly identifying an expanded facility costs 0. A random search is inefficient due to the severe class imbalance. Let $T_n$ be the number of trials until $n$ expanded facilities are found. Then $T_n=\sum_{i=1}^n T_{i-1,i}$ where $T_{i-1,i}$ is the time taken to find the $i$-th expansion after the $(i-1)$-st expansion has been found. For random search, $T_{i-1,i}$ is a geometric random variable with success probability $\frac{M-i}{N-i}$ where $M$ is total number of expansions, and $N$ the total number of images. Hence, 
\[\E[T_n] = \sum_{i=1}^n \E[T_{i-1,i}] = \sum_{i=1}^n \frac{N-i}{M-i}.\]
For us, $N=1,414$ and $M=22$. We examine how much the four models improve over this baseline cost.

\section{Results}
\label{sec:results}

We first report on the distribution of confidence values. For MLE, expansion events have a substantially higher average test statistic than non-expansions --- 23,205 versus 5,427, a difference of over 300\% ($p$-value $<0.001$ using a $t$-test). Meanwhile, none of the other three methods had statistically significant differences between the likelihoods of expansions versus no-expansions.

Next we move onto the performance metrics. Table~\ref{tab:accuracy} gives the maximum balanced accuracy, F1-score, and likelihood-size correlation of each method. MLE achieves a balanced accuracy of 78.6\%. \BCP on \NDVI is the runner-up at 56.9\%.  Due to the strong class imbalance, macro F1-scores are relatively low for all models, but at 0.60, MLE again represents a significant improvement over the other methods.
MLE achieves its maximum accuracy using a threshold of $t=5,308$, at which point it correctly classifies 86.4\% of the expansions, and 70.1\% of the non-expansions. However, many thresholds give significantly worse ratios, resulting in a low area under the PR Curve (0.1). There is hence still significant room for improvement. We emphasize though that the goal of this approach is to help \emph{prioritize} and augment enforcement resources, which is why the cost metric is particularly salient. 
With respect to the size correlation, at 0.67 ($p$-value $<0.001$) MLE is the only method which exhibits significant correlation between the likelihood and the size of the expansion. All other methods have a correlation below 0.22, none of which are statistically significant. 

\begin{table}[t]
    \centering
    \begin{tabular}{r|S[table-format=2.2, table-figures-decimal=1, table-number-alignment=left]|S[table-format=1.2, table-figures-decimal=1, table-number-alignment=left]|S[table-format=2.2, table-figures-decimal=1, table-number-alignment=left]}
         \textit{Model} & \textit{Accuracy} & \textit{F1-score} & \textit{Size Correlation} \\
         \hline 
         \BCPNDVI & 56.9\% & 0.45 & 0.21\\
         \BCPPC & 52.1\% & 0.25 & -0.20\\ 
         \BFAST & 55.2\% &  0.55& 0.00\\
         \DDMMV & 62.9\% & 0.53 & -0.02 \\
         \DDMBCP & 58.9\%& 0.50 & 0.17\\
         \textbf{MLE} & 78.6\% &0.60 & 0.67\;\;*
    \end{tabular}
    \caption{Balanced Accuracy, macro F1-scores, and Pearson correlation between confidence and size of expansion. Proposed method is in bold. In the correlation column, a star $(^*)$ indicates a $p$-value of less than 0.001.
    }
    \label{tab:accuracy}
\end{table}

\begin{figure}[h]
\centering
    \begin{minipage}{.48\textwidth}
    \subcaptionbox{\label{fig:roc}}{\includegraphics[height=4cm,left]{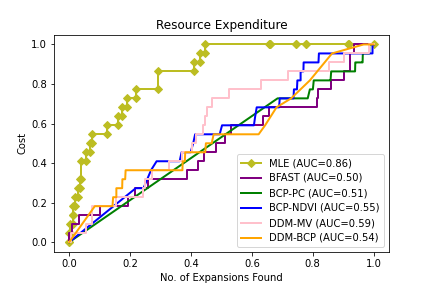}}
    \end{minipage}
    \begin{minipage}{.48\textwidth}
    \centering
    \subcaptionbox{\label{fig:cost}}{\includegraphics[height=4cm,center]{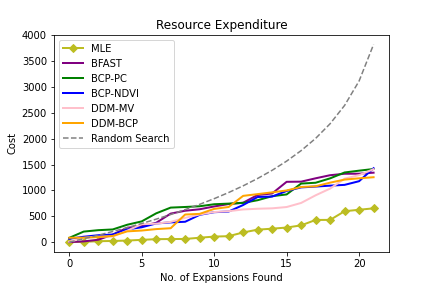}}
    \end{minipage}
    \caption{
    (a) The ROC Curve using confidence values as the prediction threshold. Locations above a given value of the test statistic are labelled as expansion. 
    (b) Reduction in cost achieved by sorting images based on their test statistic compared to random search. The $y$ axis is the cost of finding $n$ expansion events. Examining an image not exhibiting expansion is a cost of 1.   }
    \label{fig:pr_roc}
\end{figure}

Figure~\ref{fig:roc} illustrates the ROC Curves for all methods. MLE has the highest AUC at 0.86, followed by \DDMMV at 0.59, \BCPNDVI at 0.55, \DDMBCP at 0.54, \BCPPC at 0.51 and \BFAST at 0.50. Regarding cost reduction, Figure~\ref{fig:cost} demonstrates that all methods improve over random search. For each method, we rank the locations by confidence of expansion and examine the images in that order. We assume that examining a facility which did not expand has a cost of 1, and examining an expansion has a cost of 0.  A random search involves examining 3,836 false positives (facilities which did not expand) in expectation in order to find all 22 expansions. MLE examines 654 false positives, a decrease of over 80\%. The three baselines perform approximately equally, decreasing the cost by between 62\% and 64\%.  Overall, sorting by test statistic is a substantial improvement over random search. While MLE outperforms the other three methods, the savings exhibited by any of the four algorithmic approaches over an ad-hoc scan demonstrates the potential of such automated methods. 

\section{Discussion}
\label{sec:discussion}

In this work, we have demonstrated the possibility of enhancing traditional enforcement techniques with dynamic monitoring of satellite imagery. Most promising is that the methods developed herein demonstrate the possibility of near real-time environmental monitoring, using 3m/pixel resolution imagery---available at high cadence and increasingly low-cost.

Our MLE approach appears to outperform the main available baselines (\BCP, \BFAST, and \DDM) by a substantial margin. One way to think about the difference between our approach and these baselines is that our algorithm is specifically tailored to building expansion. First, \DDM (and similar algorithms, such as \textsc{CCDM}) can detect anomalies with a single image. Our MLE approach, in contrast, leverages the full time series before and after a posited changepoint.  Second, our algorithm works particularly effectively in conjunction with available building segmentation techniques.  Other approaches using \NDVI and \DDM use distinct methods (e.g., denoising, image basis) to reduce dimensionality, but, as a result, do not perform changepoint detection in a way that is as tailored to building changes. 

Because it is a lightweight approach that can be used in conjunction with segmented images, another strength is that it is not demanding computationally. On NVIDIA's Tesla v100 GPU, our method required approximately 30 seconds per location (after segmentation). Running sequentially thus yields a time of approximately twelve hours for 1,500 CAFOs. This could, of course, be sped up by running multiple GPUs in parallel.

These results are important substantively, as the gains of employing this approach would be a major improvement over the status quo of random, physical inspections. As illustrated by Figure~\ref{fig:cost}, ordering locations by their test statistic and proceeding down the list would constitute a large gain over random inspections. The U.S. Department of Agriculture estimates that there are approximately 1 million farm operations and if our base rate of expanded facilities holds, then approximately 10,000 of these facilities may be expanding each year.\footnote{This is based on a simple estimate multiplying the rate of expansion observed in our dataset ($\frac{22}{1436}$) with the number of facilities. We recognize that this simple estimate does not account for many other factors that are associated with expansion.} Augmenting conventional inspections with remote sensing can help address limitations in the resourcing of environmental protection agencies~\citep{gray2011effectiveness}.  

To illustrate, we manually searched in local records (e.g., county assessor's office, permits offices) for a sample of expanded facilities. While these records are highly decentralized and hence not easily searchable for many localities, we quickly found several facilities that were classified as ``vacant land,'' both before and after significant expansion events.  The comparison is stark: comparable permitted facilities can pay over 50 times the property tax.  With access to permit records, our method could hence easily scale the approach pioneered by Massachusetts, which manually compared satellite imagery against permit records to identify illegal wetland modification \citep{clayton_wetlands_2004}. 

We close with several thoughts on directions for future research. First, the likelihood-based expansion technique relies heavily on the quality of the segmentation model. Developing better CAFO-specific segmentation models would thus help improve the results. Second, with a larger labeled set of images, it would be worthwhile comparing the performance of end-to-end deep learning approaches to changepoint detection.  Third, it would be worth investigating generalizations of the likelihood approach.  Equation~\eqref{eq:transition} can be modified and researchers may wish to incorporate other information about CAFOs, such as proximity of the added building, conformity with direction of existing sheds, and closures of facilities. Fourth, it could prove beneficial to combine the Dynamic Detection Model (\DDM) with the segmenter, as opposed to running it on the image bands. Doing so might reduce the number of false positives arising from abrupt changes in the surrounding landscape (see Appendix~\ref{app:figs}).

Last, as emphasized in the introduction, nothing about this approach is specific to CAFOs besides the initial segmentation model. By adapting this model, the likelihood based approach could be used to detect any structural changes occurring over time. It would be fruitful to apply this method to the range of other environmental challenges, such as zoning violations, habitat modification, and deforestation. 

We hope to have provided a useful approach that leverages rapid increases in the availability of satellite imagery to enable remote sensing to provide  actionable insights for real-time environmental enforcement. Doing so could dramatically improve the allocation of scarce regulatory resources to where most needed.

\bibliographystyle{elsarticle-harv}
\bibliography{references}

\clearpage
\appendix 

\section{Sensitivity to Segmentation}
\label{app:segment}

One of the contributions of our approach is a lightweight changepoint detection method that does not rely on the specific implementation of a segmentation algorithm. We believe this to be a strength of this contribution, as researchers can choose any preferred segmentation approach.  We nonetheless examine the effect of segmentation accuracy on the results of the MLE approach. 

We artificially degrade the quality (pixel accuracy) of the segmenter by 1\%, 2\% and 5\%, and study effects on the MLE approach based on such degraded segmentation. These segmentation differences are representative of differences between state-of-the-art approaches~\citep{ahmed2020comparison, cai2021comparative, jiwani2021semantic}. 

Our findings from this ablation study are twofold.  First, as expected, reduction in segmentation accuracy is associated with performance degradation in the MLE approach.
Second, this performance degradation is relatively small. For instance, MLE accuracy with the U-Net segmented images yields 78.6\% accuracy, which drops to 78.3\% with 1\% noise, 78.2\%  with 2\% and 78.1\% with 5\%. We conclude that the MLE approach is likely to remain valuable for change detection of buildings across a range of current segmentation approaches. 
\newpage

\clearpage
\section{Illustration of MLE and \textsc{DDM}}
\label{app:figs}

\begin{figure}[h!]
    \begin{minipage}{0.48\textwidth}
    \includegraphics[scale=0.3,center]{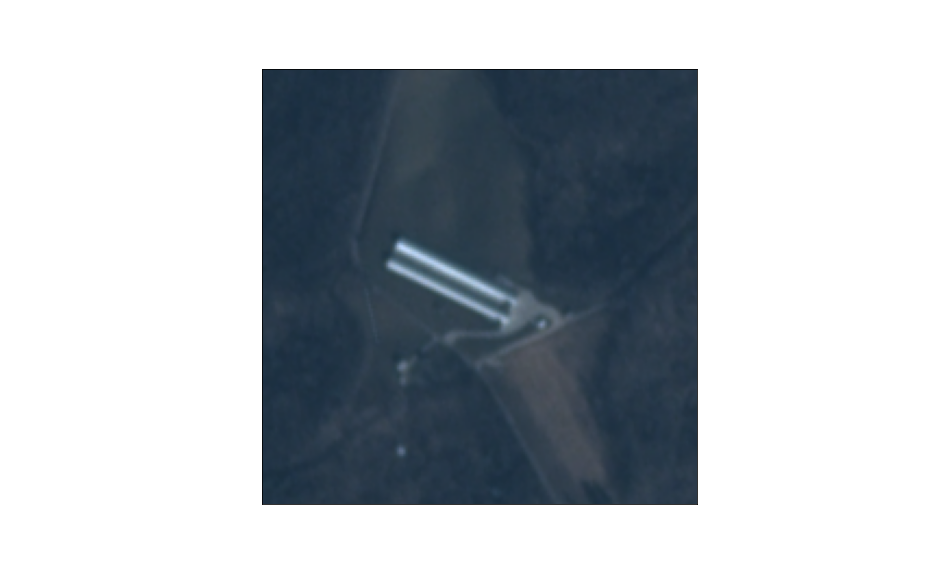}
    \vspace{-1cm}
    \subcaption{}
    \end{minipage}
    \hfill
    \begin{minipage}{0.48\textwidth}
    \includegraphics[scale=0.3,center]{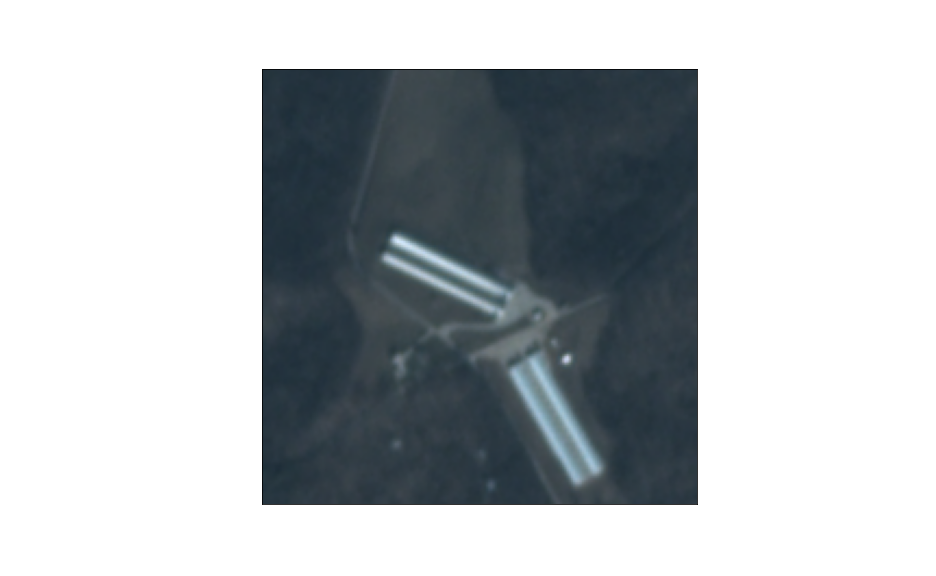}
    \vspace{-1cm}
    \subcaption{}
    \end{minipage}
    \begin{minipage}{0.48\textwidth}
    \includegraphics[scale=0.3,center]{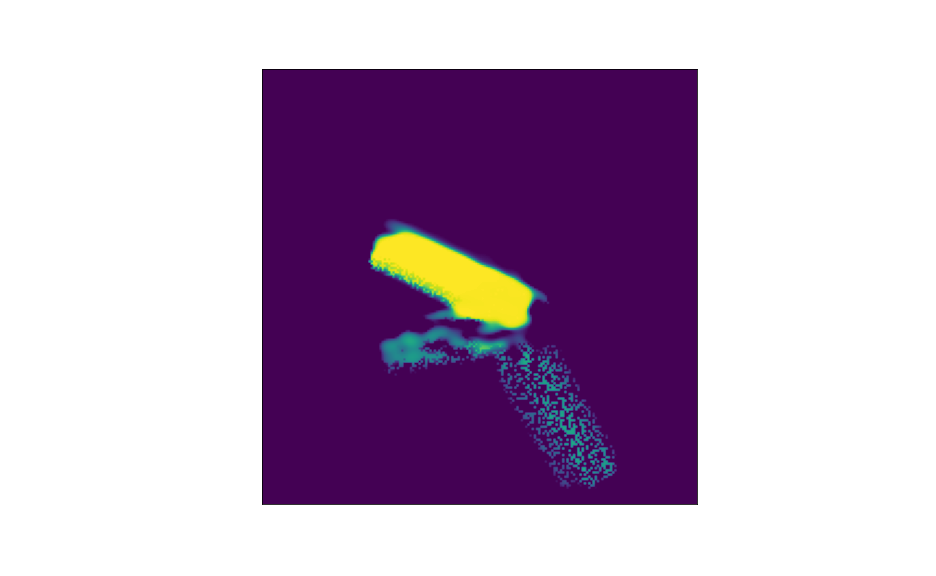}
    \vspace{-1cm}
    \subcaption{}
    \end{minipage}
    \hfill
    \begin{minipage}{0.48\textwidth}
    \includegraphics[scale=0.3,center]{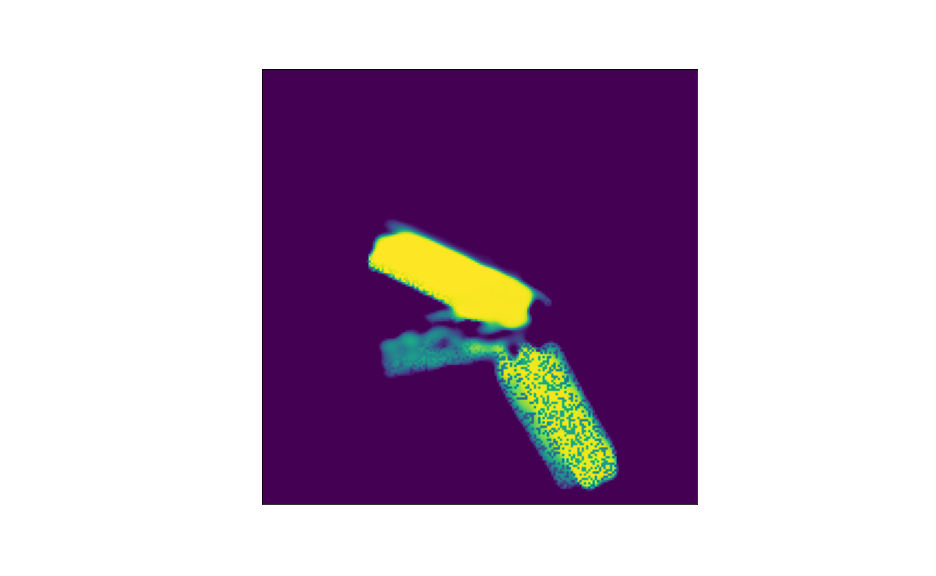}
    \vspace{-1cm}
    \subcaption{}
    \end{minipage}
    \begin{minipage}{0.32\textwidth}
    \includegraphics[scale=0.25,center]{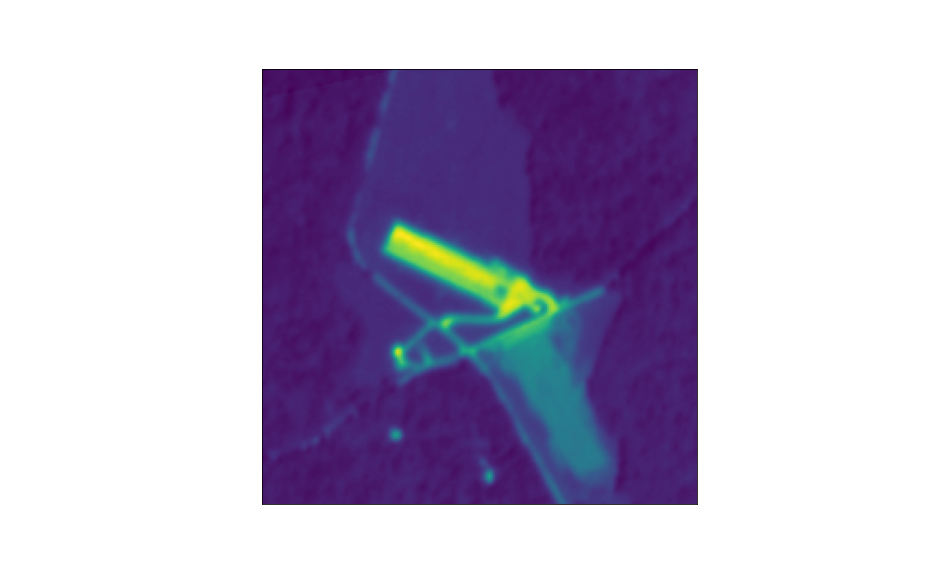}
    \vspace{-1cm}
    \subcaption{Predicted}
    \end{minipage}
    \begin{minipage}{0.32\textwidth}
    \includegraphics[scale=0.25,center]{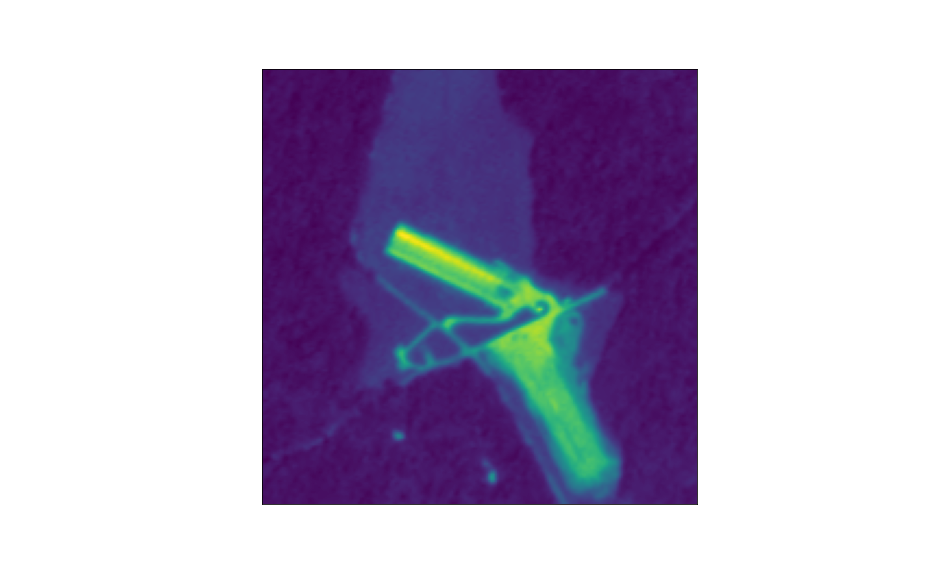}
    \vspace{-1cm}
    \subcaption{Truth}
    \end{minipage}
    \begin{minipage}{0.32\textwidth}
    \includegraphics[scale=0.25,center]{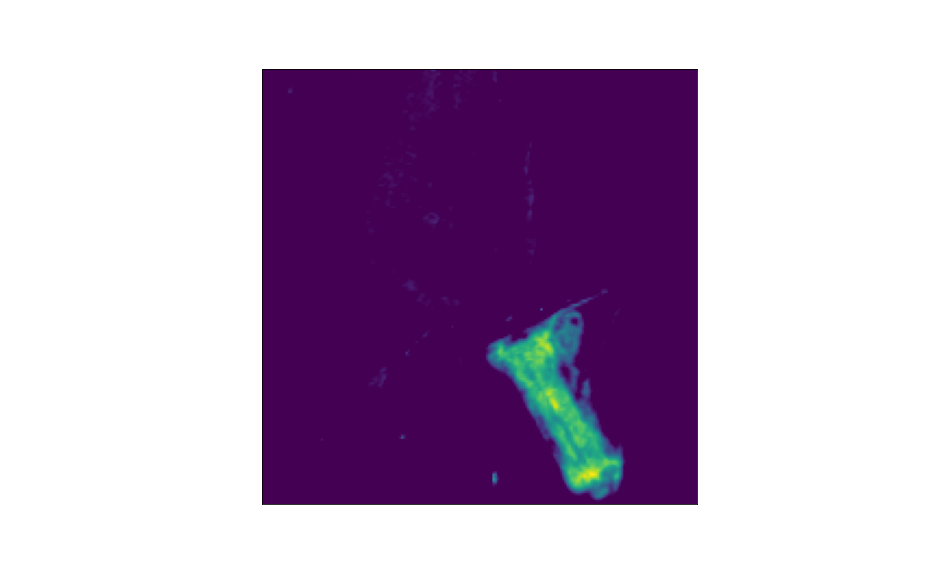}
    \vspace{-1cm}
    \subcaption{Anomaly}
    \end{minipage}
    \caption{(a) and (b): images of an expanded CAFO at the beginning and end of the year. The expansion occurred during week 28. (c) and (d): the pre and post expansion model generated by MLE. (e), (f) and (g): The predicted green (G) image band at week 28 (based on a weighted average of the previous ten weekly images), the true band at week 28, and the resulting anomaly score $Z_{ij,G}^{28}$ between the two.}
    \label{fig:both_good}
\end{figure}

In this Appendix, we illustrate how MLE and \DDM perform in our setting. Figure~\ref{fig:both_good} displays a setting in which both methods perform well at detecting an added barn. The top panel plots raw images before and after an expansion event. The middle panels plot the pre- and post-expansion models (i.e., $\F^0$ and $\F^+$) generated by our MLE approach, showing that the MLE approach focuses on the added building. 

The bottom panel illustrates the \DDM model: the left image plots the predicted band immediately prior to an expansion event, based on a weighted sum of the of the previous 10 weekly images; the middle panel plots the actual band, including the expansion event; and the right panel plots the pixel-wise $z$-score between the true and predicted images.  This \DDM result also shows that the approach reasonably focuses on expanded barn location, confirming that this is a reasonable implementation of the \DDM baseline.  

\begin{figure}[th]
    \centering
    \begin{minipage}{0.48\textwidth}
    \includegraphics[scale=0.3,center]{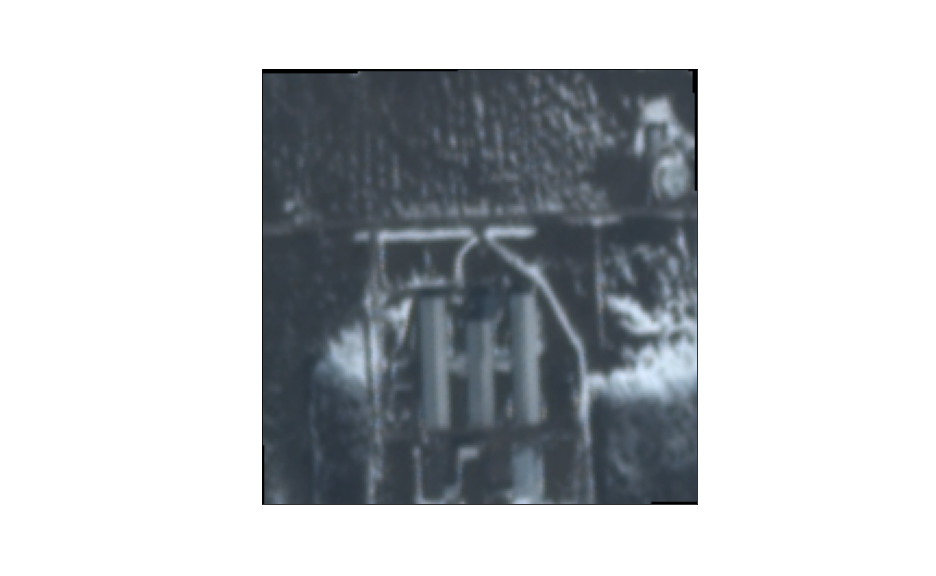}
    \vspace{-1cm}
    \subcaption{}
    \end{minipage}
    \hfill
    \begin{minipage}{0.48\textwidth}
    \includegraphics[scale=0.3,center]{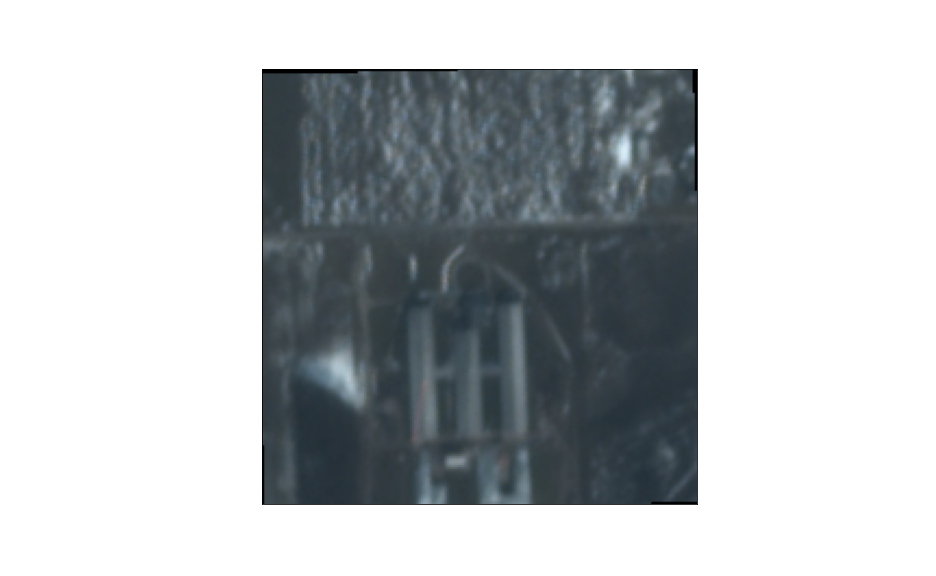}
    \vspace{-1cm}
    \subcaption{}
    \end{minipage}
    \begin{minipage}{0.48\textwidth}
    \includegraphics[scale=0.3,center]{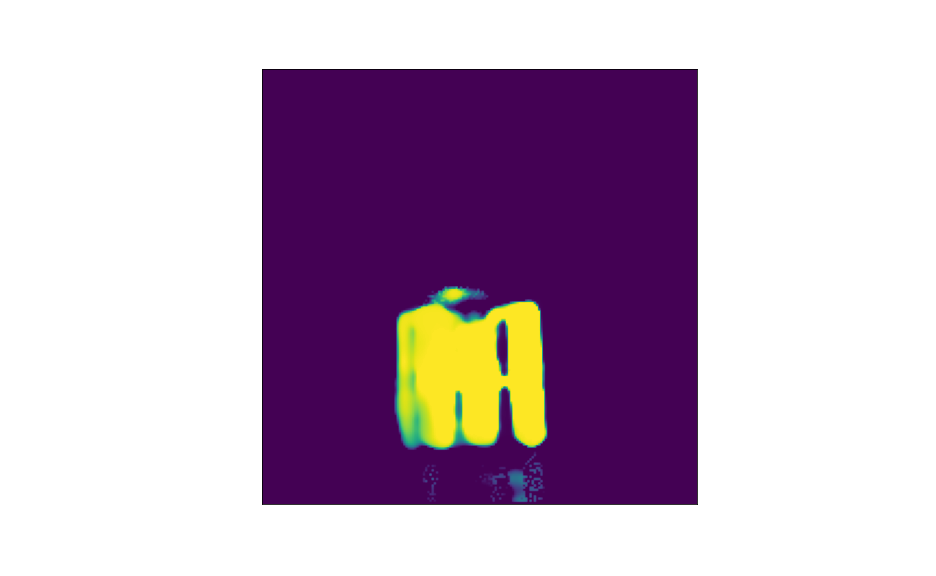}
    \vspace{-1cm}
    \subcaption{}
    \end{minipage}
    \hfill
    \begin{minipage}{0.48\textwidth}
    \includegraphics[scale=0.3,center]{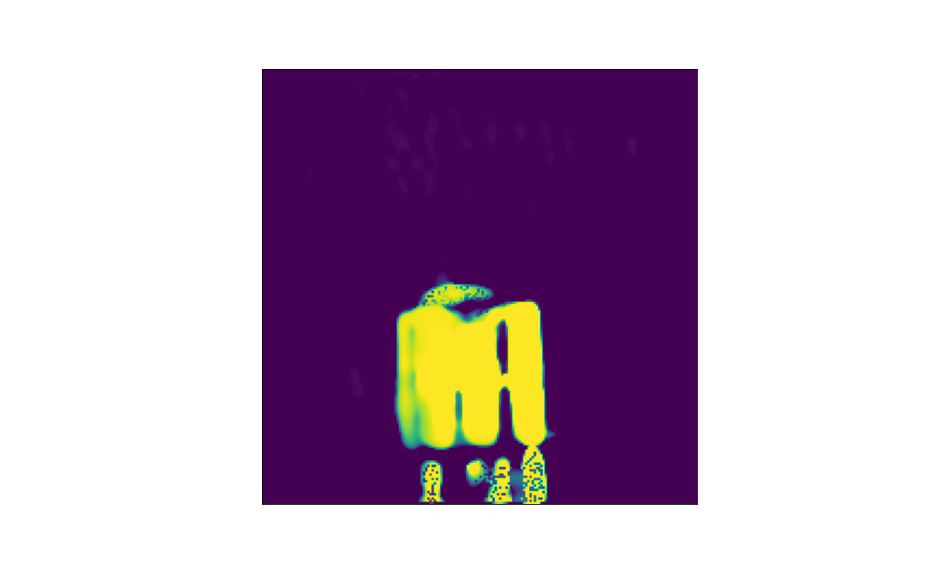}
    \vspace{-1cm}
    \subcaption{}
    \end{minipage}
    \begin{minipage}{0.32\textwidth}
    \includegraphics[scale=0.25,center]{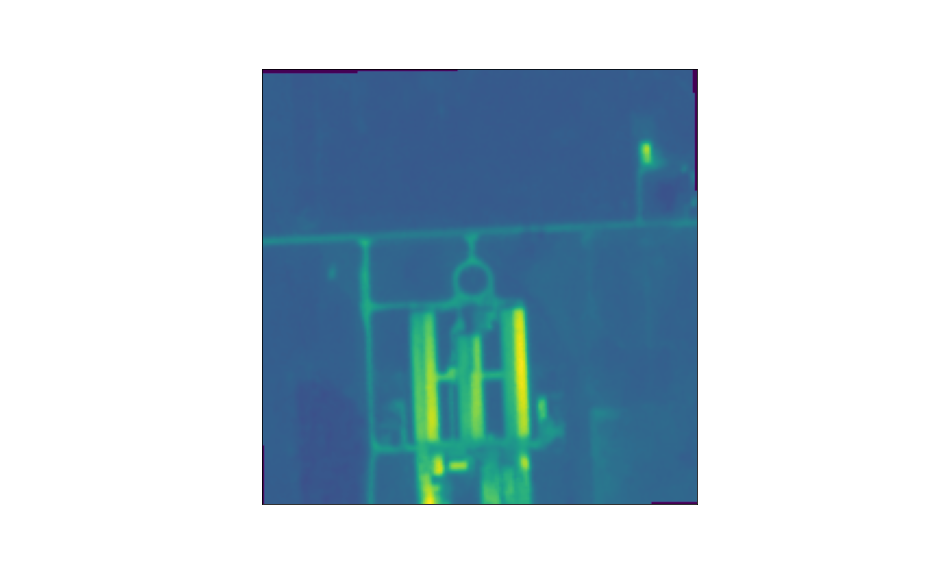}
    \vspace{-1cm}
    \subcaption{Predicted}
    \end{minipage}
    \begin{minipage}{0.32\textwidth}
    \includegraphics[scale=0.25,center]{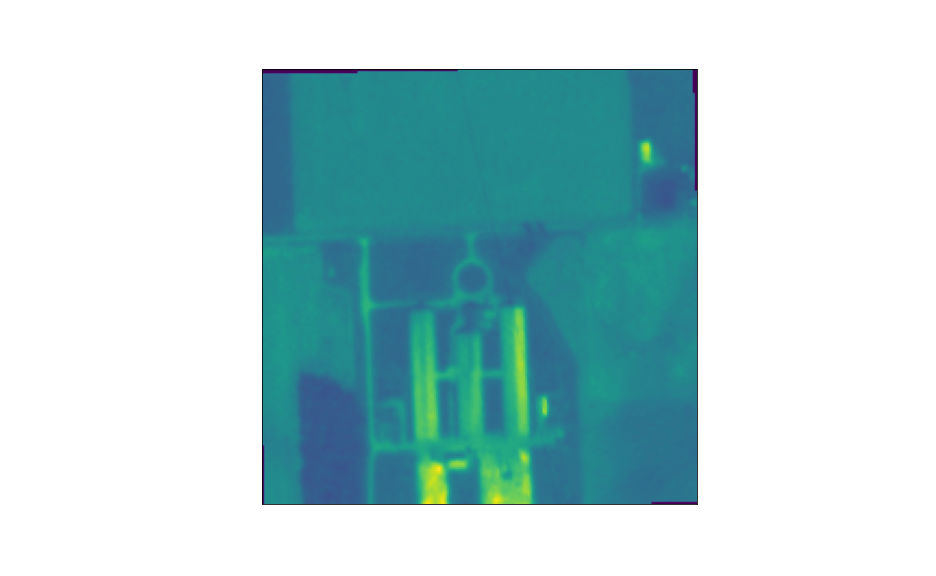}
    \vspace{-1cm}
    \subcaption{Truth}
    \end{minipage}
    \begin{minipage}{0.32\textwidth}
    \includegraphics[scale=0.25,center]{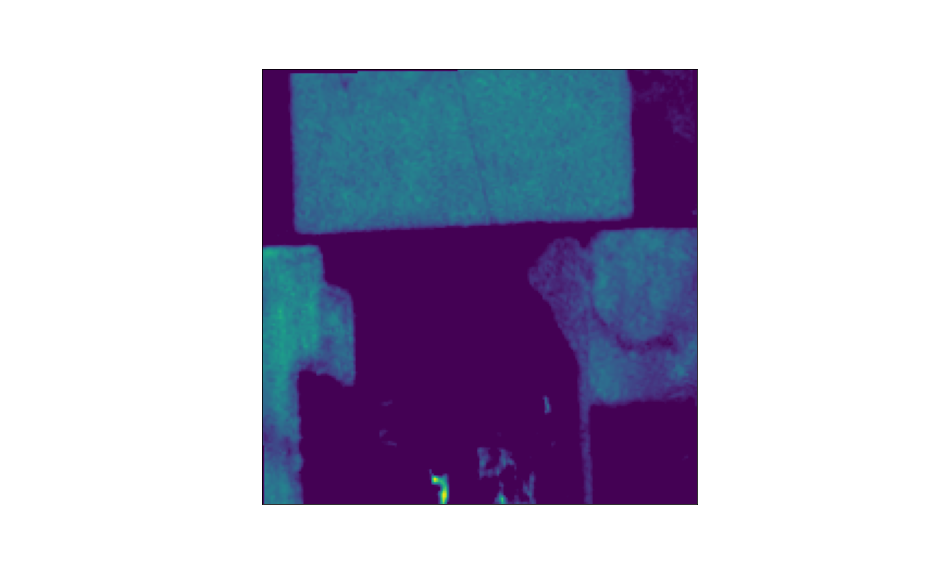}
    \vspace{-1cm}
    \subcaption{Anomaly}
    \end{minipage}
    \caption{(a) and (b): images of a CAFO at the beginning and end of the year. Note that there was no expansion. (c) and (d): the pre and post expansion model generated by MLE. Because there was no expansion, the pre and post models are very similar, the differences being due to segmentation noise. (e), (f) and (g): The predicted green (G) image band at week 40 (based on a weighted average of the previous ten weekly images), the true band at week 40, and the resulting anomaly score $Z_{ij,G}^{40}$ between the two. Due to abrupt changes in the surrounding fields, there are many pixels with significant anomaly scores, thus resulting in a false positive for \DDM.}
    \label{fig:ddm_bad}
\end{figure}

Figure~\ref{fig:ddm_bad} illustrates a non-expansion event in which the MLE model is relatively stable, but the \DDM approach yields a false positive due to abrupt changes in the surrounding landscape. This shows a drawback of \DDM: drastic changes of any kind can be interpreted to be examples of building expansion. As noted in the Discussion, this kind of anomaly detection may underperform relative to the MLE approach because (a) it does not take advantage of the full time series before and after an expansion event and/or (b) does not focus the change point detection on buildings using  segmentation. A potential direction for future work may be to combine \DDM with the segmentation approach. 

\end{document}